\documentclass[letterpaper, 10 pt, conference]{ieeeconf}

\IEEEoverridecommandlockouts                              

\usepackage{graphicx}
\usepackage{bbm}
\usepackage{mathtools}
\usepackage[colorinlistoftodos, textwidth=2cm, textsize=tiny]{todonotes}

\newcommand{\Prob} {\ensuremath \mathbf{P}  }
\title{\LARGE \bf
Feature discovery and visualization of robot mission data using convolutional autoencoders and Bayesian nonparametric topic models
}
\author{Genevieve Flaspohler$^{1, 2}$, Nicholas Roy$^{1}$, and Yogesh Girdhar$^{2}$
\thanks{$^{1}$Computer Science and Artificial Intelligence Laboratory (CSAIL), Massachusetts Institute of Technology, Cambridge MA, 02139, USA {\tt\small \{geflaspo, nickroy\}@mit.edu}}%
\thanks{$^{2}$Department of Applied Ocean Physics and Engineering, Woods Hole
Oceanographic Institution, Woods Hole MA, 02543, USA {\tt\small yogi@whoi.edu}}%
}
\begin{document}

\maketitle
\thispagestyle{empty}
\pagestyle{empty}

\begin{abstract}
The gap between our ability to collect interesting data and our ability to analyze these data is growing at an unprecedented rate. Recent algorithmic attempts to fill this gap have employed unsupervised tools to discover structure in data. Some of the most successful approaches have used probabilistic models to uncover latent thematic structure in discrete data. Despite the success of these models on textual data, they have not generalized as well to image data, in part because of the spatial and temporal structure that may exist in an image stream.
    
We introduce a novel unsupervised machine learning framework that incorporates the ability of convolutional autoencoders to discover features from images that directly encode spatial information, within a Bayesian nonparametric topic model that discovers meaningful latent patterns within discrete data. By using this hybrid framework, we overcome the fundamental dependency of traditional topic models on rigidly hand-coded data representations, while simultaneously encoding spatial dependency in our topics without adding model complexity. We apply this model to the motivating application of high-level scene understanding and mission summarization for exploratory marine robots. Our experiments on a seafloor dataset collected by a marine robot show that the proposed hybrid framework outperforms current state-of-the-art approaches on the task of unsupervised seafloor terrain characterization. 
\end{abstract}

\section{Introduction}
The benthic deep sea, the largest two-dimensional habitat on earth, is difficult to study and vastly unexplored. Autonomous underwater vehicles (AUVs) are filling observational gaps by collecting large datasets consisting of multiple sensor modalities, including seafloor imagery.  This paper presents a novel unsupervised machine learning technique to discover and visualize structure in image datasets, enabling concise mission summarization and equipping exploratory robots with the capacity to describe their environment semantically, a precursor to adaptive real-time exploration.  Although the focus of this work is the underwater domain, the proposed approach is applicable to any domains where there exist large volumes of unstructured image sequence data that would typically require manual analysis, such as remote sensing and long term monitoring. 


Some of the most successful models for discovering structure within discrete data without supervision are Bayesian topic models, such as the latent Dirichlet allocation (LDA) \cite{Blei2003} and its non-parametric extension, the Hierarchical Dirichlet process (HDP) \cite{Teh2010}. Initially applied to text corpora, the modeling assumptions made by LDA and HDP allow them to discover useful latent structure that often corresponds to cohesive, human-understandable topics~\cite{Chang2009}. This property of topic models led to impressive results in areas such as text clustering \cite{Blei2003}, corpora summarization, and recommender systems \cite{Krestel}.  

The success of Bayesian topic models for semantic understanding of text documents led to their adaptation to the computer vision domain. By replacing textural words with discrete visual features, LDA and HDP models can be applied directly to image data \cite{Bosch:2006, FeiFei:2005:CVPR, Wang2007}. The most popular discretization of an image into visual words employs standard features such as SIFT \cite{Lowe:IJCV:2004}, SURF \cite{Bay:ECCV:2006}, or Oriented BRIEF (ORB) \cite{RubleeE2011}. Visual topic models have been used successfully in robotics applications for for unsupervised scene understanding \cite{Steinbergb} and adaptive mission planning \cite{Girdhar2015}. 


\begin{figure}[t]
    \centering
    \includegraphics[width=\columnwidth]{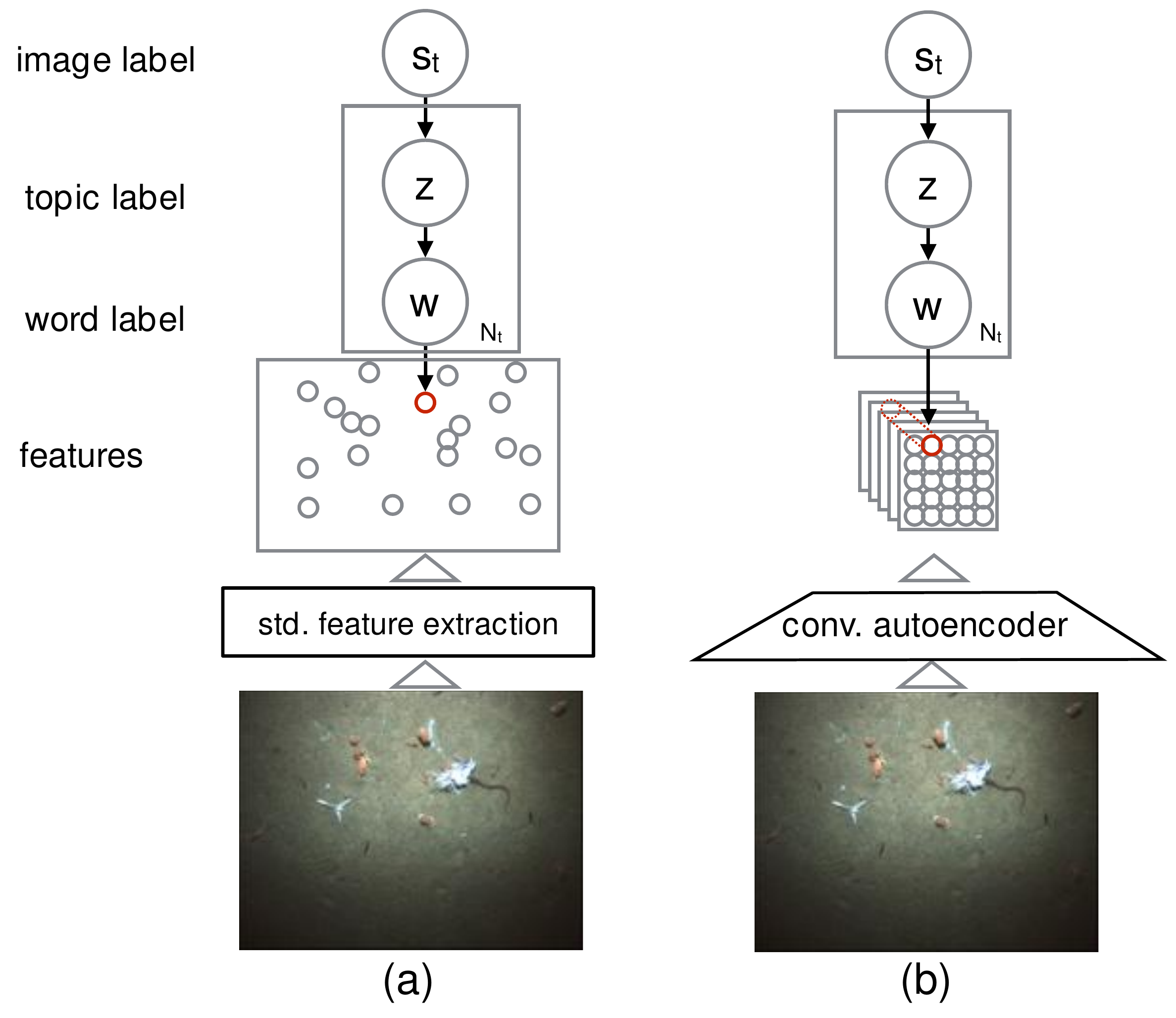}
    \caption{Two scene modeling techniques evaluated in this paper. Here $s_t$ is the image label, and $z$ is the topic label of a visual word $w$ in the input image. (a) A baseline spatiotemporal topic model using standard computer vision features as input. (b) The proposed  spatio-temporal topic model using convolutional autoencoder-based features.}
    \label{fig:overview}
    \vspace{-.105in} 
\end{figure}
    
However, the modeling capacity of topic models is fundamentally limited by the expressive power of the observed `words'. Hand-crafted image features capture low-level patterns based on local image gradients. In contrast, deep neural models are often able to learn more complex, domain-specific features. Several papers have leveraged this property of neural networks to build more expressive models of textual data \cite{Mikolov2012}, \cite{Shen2014}.  In this paper, we make the natural extension to unsupervised feature discovery for image data. 

Much like textual data, image data show strong spatial correlations. These correlations are ignored by the simplifying bag of words (BOW) assumption made in most Bayesian topic models. Ideally, data features could encode these spatial correlations directly. Convolutional autoencoders (CAE) \cite{Masci} preserve spatial relationships in data and hence are a powerful method for discovering useful features for image data. However, these features have not yet been incorporated into a topic modeling framework, in part because of the challenges of designing a CAE network architecture that produces useful features within a topic modeling context. 


This work presents a CAE architecture that discovers feature representations directly from an image dataset and applies those features within an HDP-based topic modeling framework to discover cohesive visual topics.  We explore the performance of this hybrid HDP-CAE model on the motivating application of autonomy and mission summarization for exploratory marine robots. We evaluate how well the topics discovered by the hybrid HDP-CAE model correspond to biologically distinct seafloor terrains and compare the hybrid HDP-CAE model to an HDP model using standard image features. Finally, we quantify the performance of the hybrid model on the secondary task of identifying anomalous images within a dataset. The probabilistic anomaly detection enabled by Bayesian topic models can inform more effective mission planning for exploratory marine robots and is more broadly useful for data summarization and visualization. 

All models are evaluated on a realistic dataset that an individual biologist or data scientist could collect and wish to analyze, consisting of less than 4,000 seafloor images collected in-situ by a marine robot. Even in this small-data domain, we demonstrate that state-of-the-art performance can be achieved by applying neural feature discovery and nonparametric topic modeling to the task of unsupervised seafloor terrain characterization.


\begin{figure*}[thb!]
    \centering
    \includegraphics[scale=0.45]{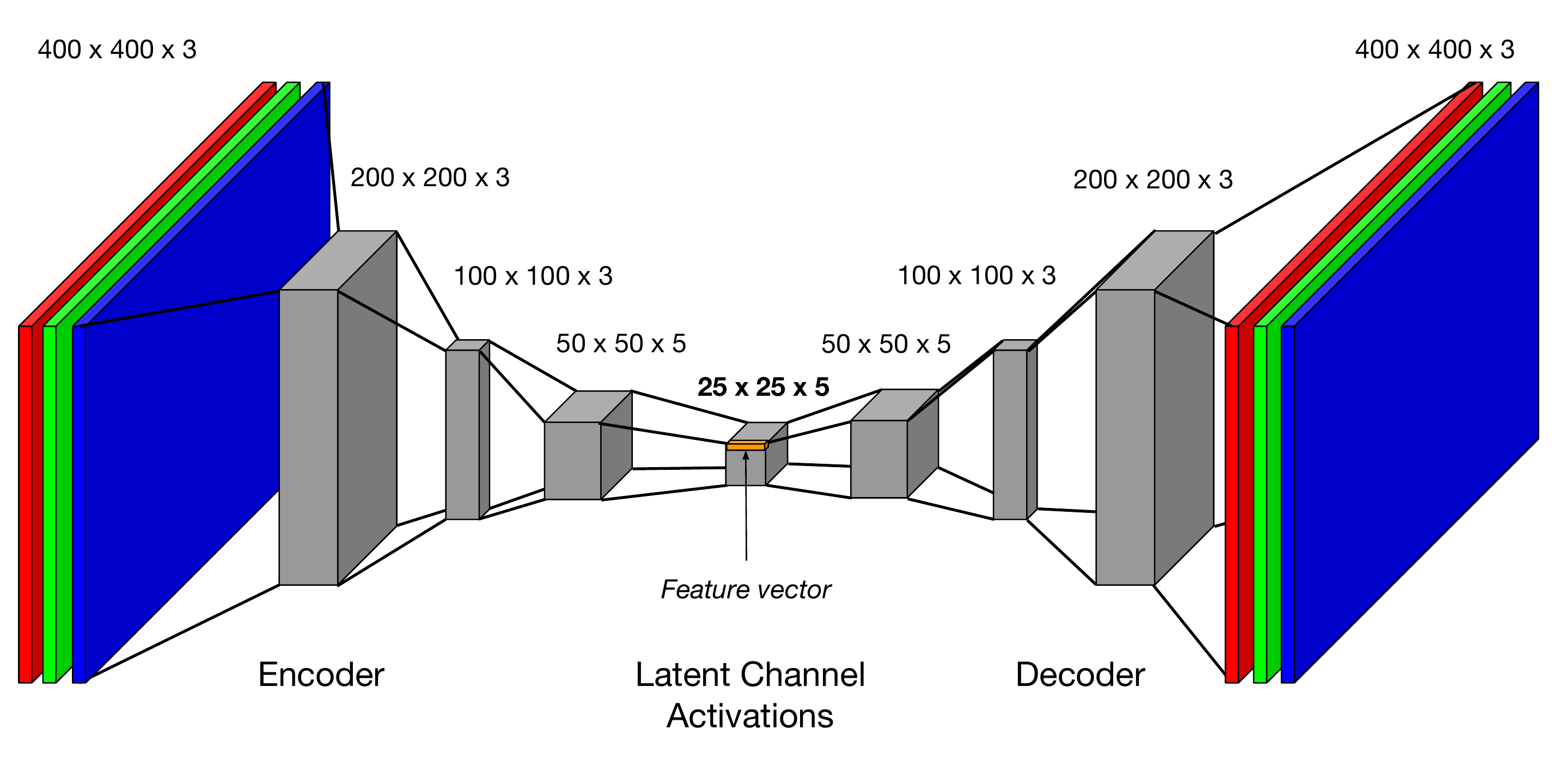}
    \caption{Network architecture for the convolutional autoencoder (CAE) used to extract low level visual features from the image datasets. Network specific parameters were set as: training epochs (400 epochs), output channels (ordered by encoding layer 3-3-3-5-5 channels), stride (2), and convolutional filter size (ordered by encoding layer 10-10-3-3 pixels),}
    \label{fig:arch}
\end{figure*}

\section{Related Work}
Recent efforts have leveraged the power of neural models to discover data features within the context of topic modeling. Many, however, continue to rely on predefined data features at some level; frameworks that overcome this dependence have difficulty incorporating custom features within a completely unsupervised Bayesian topic model.

For textual data, Mikolvo et al. \cite{Mikolov2012} propose the reverse of the architecture we present here; an LDA model is used to produce a contextual feature vector that is input into a recurrent neural network for contextually-aware language modeling. While powerful, this model does not address the model's dependence on standard word features and employs a BOW assumption. In \cite{Shen2014}, the BOW assumption is relaxed. Instead, a convolution operation maps variable length text sequences into a low-dimensional latent space. Unlike the work presented here, simple distance-based clustering is applied to discover semantically similar documents in place of a Bayesian topic model.
     
For image data, Wan et al. \cite{Wan} introduce a hybrid neural-Bayesian topic model based on a Deep Boltzmann Machine (DBM). As in this work, the feature representation discovered by the DBM is fed directly into an HDP topic model to discover visual topics.  However, instead of discovering features directly from image data, SIFT features are extracted from the image and  the neural network learns an image representation based on these features, thereby not reducing the dependence on human-designed features. 

The work most similar to our own is presented in \cite{Salakhutdinov}.  The Hierarchical-Deep model introduced uses an HDP to learn priors over the activations of a DBM. In this way, the model is able to learn generic features from image data that enable learning image classes from very few examples. Each image in the model is annotated with a lower level class and the HDP discovers a hierarchy over these low-level classes. While suitable for the goal of one-shot learning, this supervision limits the generality of the Hierarchical-Deep model to a purely unsupervised problem. The model is also not convolutional, limiting the utility of the learned features. Convolutional autoencoders are directly able to model spatial correlations in image data and therefore are more suited to discover useful image representations.

Additionally, none of the aforementioned works evaluate their models on the small or medium-sized datasets that are prevalent in unsupervised learning applications. Instead, they use large (4 million+) standard image datasets \cite{Salakhutdinov} or 2D toy, simulated images \cite{Wan}. 

Other works have incorporated neural feature learning for robotics applications outside of a topic modeling context. Naseer et al. \cite{Naseer} use up-convolutional networks to discover latent feature representations for the task of segmenting images. Rao et. al \cite{Rao2014} use an autoencoder to learn features for classification of marine images. However, both of these works require human annotation and thus are not applicable in an unsupervised setting. 
\section{Methods}
\label{sec:approach}
 In the following sections, we provide a brief review of topic models and then discuss the two major components of the proposed hybrid HDP-CAE model: 1) a spatio-temporal HDP topic model and 2) a pipeline for discretizing an image into visual words using a convolutional autoencoder. 

\subsection{Bayesian topic models}
Topic models \cite{Blei2003, Griffiths:2004} seek to uncover semantic structure in a corpus of discrete data, segmented into documents. Topic models propose that each observed word in a document is generated by a latent topic and each document in a corpus has its own probability distribution over topics.  Using word co-occurrences and distribution sparsity constraints, the distribution over topics $z_i$ for each word $w_i$ can be inferred. Under this model, the probability of the $i$th word $w_i$ can be written as: 

\begin{equation}
    \Prob(w_i) = \sum_{k=1}^K \Prob(w_i| z_i = k) \Prob(z_i = k | d) 
\end{equation}
where $K$ is total number of topics, $\Prob(w_i | z = k)$ is the probability of word $i$ under topic $k$, and $\Prob(z_i = k | d)$ is the probability of topic $z_i$ in the current document $d$.

\subsection{Realtime spatio-temporal HDP model}
\label{sec:hdp}
Traditional topic modeling frameworks treat each word in a document as exchangeable. We instead adapt the ROST HDP model presented in \cite{Girdhar2015}, which relaxes the BOW assumption and explicitly models the correlation between spatio-temporal neighborhoods in a continuous image stream. ROST uses a Dirichlet process to model the growth in number of topics with the size and complexity of the data. A biased Gibbs sampler \cite{Girdhar2015Gibbs} enables online computation of the posterior distribution over topics for observed visual words.
    
The ROST model factors the probability of observing the visual word $w_i$ at location $x$ and time $t$ in terms of the topic label variables $z_i$. 

\begin{eqnarray}
\Prob(w_i| x,t ) = \sum_{k=1}^K \Prob(w_i | z_i = k) \Prob(z_i = k | x,t).
\end{eqnarray}
    
The distribution $\Prob(w | z=k)$ is invariant to the spatio-temporal location of the observation, while $\Prob(z=k | x,t)$ models the distribution of topic labels in the spatio-temporal neighborhood of location $(x,t)$. $K$ is total number of topics that have at least at least one or more words assigned to them, plus one more to encode the possibility of creating a new topic for word $w_i$. ROST uses the Dirichlet distribution to model $\Prob(w|z)$, allowing for control of the sparseness of the topic model, whereas $\Prob(z|x,t)$ is modeled using the Chinese Restaurant Process \cite{Teh:2006:HDP}, removing the need to predetermine the total number of unique topics.  
\subsection{Convolutional autoencoder architecture and training} 
\label{sec:cae}
To extract a discrete list of words from an image, we exploit the ability of neural models to discover useful abstract representations of data without supervision. We train a CAE architecture following the encoder/decoder paradigm described in \cite{Masci}. The input image is first transformed into a lower dimensional bottleneck layer using successive convolution operations and rectified linear unit (ReLU) activations and then expanded back to its original size using a deconvolution operation with tied weight matrices. We call the channels in the bottleneck layer the latent channel activations (LCA). 

The squared error between the original image and the image reconstruction provides an unsupervised loss function which allows the weight matrices for each layer to be learned. The network is trained using stochastic gradient descent with L2 regularization on the weight matrices. Because our goal is to treat the nodes in the bottleneck layer as non-overlapping features of the image, the neuron redundancy encouraged by dropout regularization is actually undesirable \cite{Srivastava2014}, so we do not include dropout. In this unsupervised setting, the entire dataset is used to train and test the model, so generalization is much less of a concern than in supervised learning problems. In this paper, all models are also trained without max-pooling/unpooling layers; dimensionality reduction is achieved using a stride greater than one and overlapping convolutional filter windows. During experimentation, we found that the inclusion of max-pooling and unpooling layers decreased the expressive power of the LCA, contrary to \cite{Masci}, so the final models were purely convolutional with ReLU nonlinearity. 
    
The CAE network architecture used to produce the results in this paper is shown in Figure \ref{fig:arch}. The network consists of four encoding layers and four associated decoding layers.  Each sequential encoding layer increases the number of output channels while decreasing the height and width of each individual channel, following a pyramid architecture. The architecture-specific parameters, such as number of training epochs (400 epochs), output channels (ordered by encoding layer 3-3-3-5-5 channels), stride (2), and convolutional filter size (ordered by encoding layer 10-10-3-3 pixels), were determined empirically.  After training, we remove the decoding layers of the network and use the LCA to generate low dimensional image features.  The Tensorflow \cite{Abadi} implementation is available online\footnote{ http://warp.whoi.edu/code/}.

\subsection{Generating a visual vocabulary for topic models}
HDPs require discrete data drawn from a vocabulary $\cal V$. To produce a vocabulary for CAE features, the CAE is trained on several example ocean mission datasets and the LCA for each image are extracted as described in Section \ref{sec:cae}. This $25 \times 25 \times 5$ tensor is segmented into $625$ feature vectors of length $5$ by taking slices across LCA channels, as shown in Figure \ref{fig:arch}. These features are then clustered using the $k$-means algorithm into $|\cal V|$ clusters, where $|\cal V|$ is the desired vocabulary size. The centroid of each cluster represents a visual vocabulary word. Because the LCA are low dimensional ($ 5 \times 1$ pixels using the architecture in Figure \ref{fig:arch}), as compared to 128-dimensional SIFT/SURF features, this clustering is relatively efficient.

Given a new image, visual features are extracted and mapped to the visual word $v_i \in \cal V$ corresponding to the nearest neighbor in the space of cluster centroids.


\subsection{CAE feature visualization}
\label{sec:raw-cae}
The LCA discovered by our CAE model correspond to a low-dimensional, abstract representation of the image. In later sections we will apply these features within a topic modeling framework and attempt to visualize the properties of the latent channels constructed in this manner.

To quantify the strength of a latent channel's response to a particular input image, we consider the magnitude of each latent channel at a particular pixel location $p_{ij}$ in the LCA.  For each of the $5 \times 5$ pixels, we assign the pixel $p_{ij}$ to the channel with the maximum activation at location $(i, j)$. The magnitude of a latent channel M is equal to number of pixels for which it had the maximal value. This approch produces a clearer segmentation between channels than directly plotting channel magnitudes. Different channels have different baseline activations. To compare between channels, we normalize each channel's activation between their minimum value and maximum value before plotting. 




\begin{figure*}[thb!]
\centering
\includegraphics[scale=0.78]{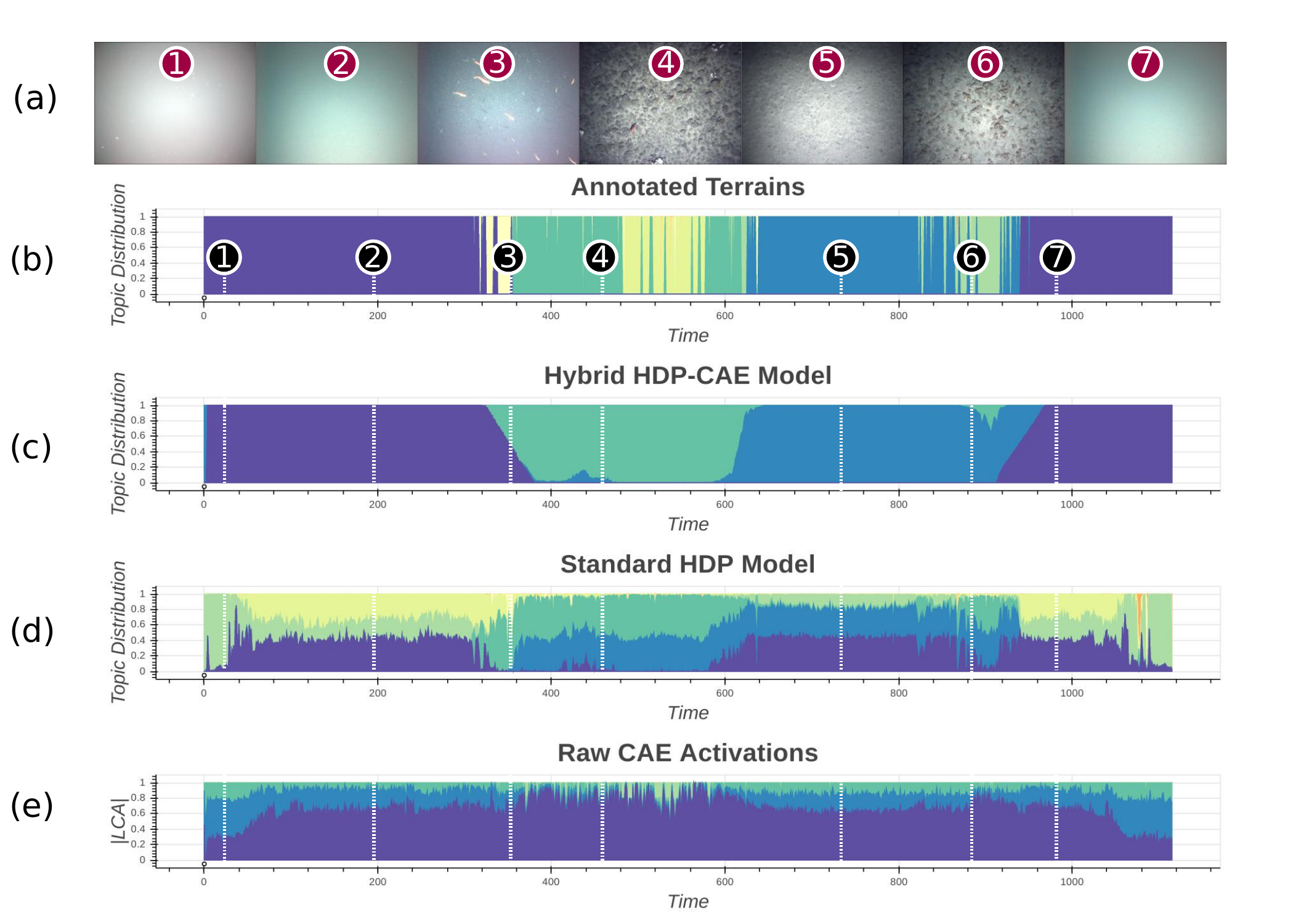}
    \caption{Results for unsupervised topic models versus hand-annotated terrain labels (b) for the Mission I dataset. Example images from the dataset are shown in (a). To generate plots (c,d), visual words are extracted from an image at time $t$ and assigned a topic label $z_i$ as described in the text. The proportion of words in the image at time $t$ assigned to each topic label is shown on the y-axis, where different topics are represented by colors. Colors are unrelated across plots.
The hybrid HDP-CAE model (c), using more abstract features, is able to define topics that correspond more directly to useful visual phenomena than the HDP model using standard image features (d). The learned feature representation is visualized as described in the text in (e).}
\label{fig:missI}
\vspace{-.105in} 
\end{figure*}
\begin{figure*}[thb!]
\centering
\includegraphics[scale=0.76]{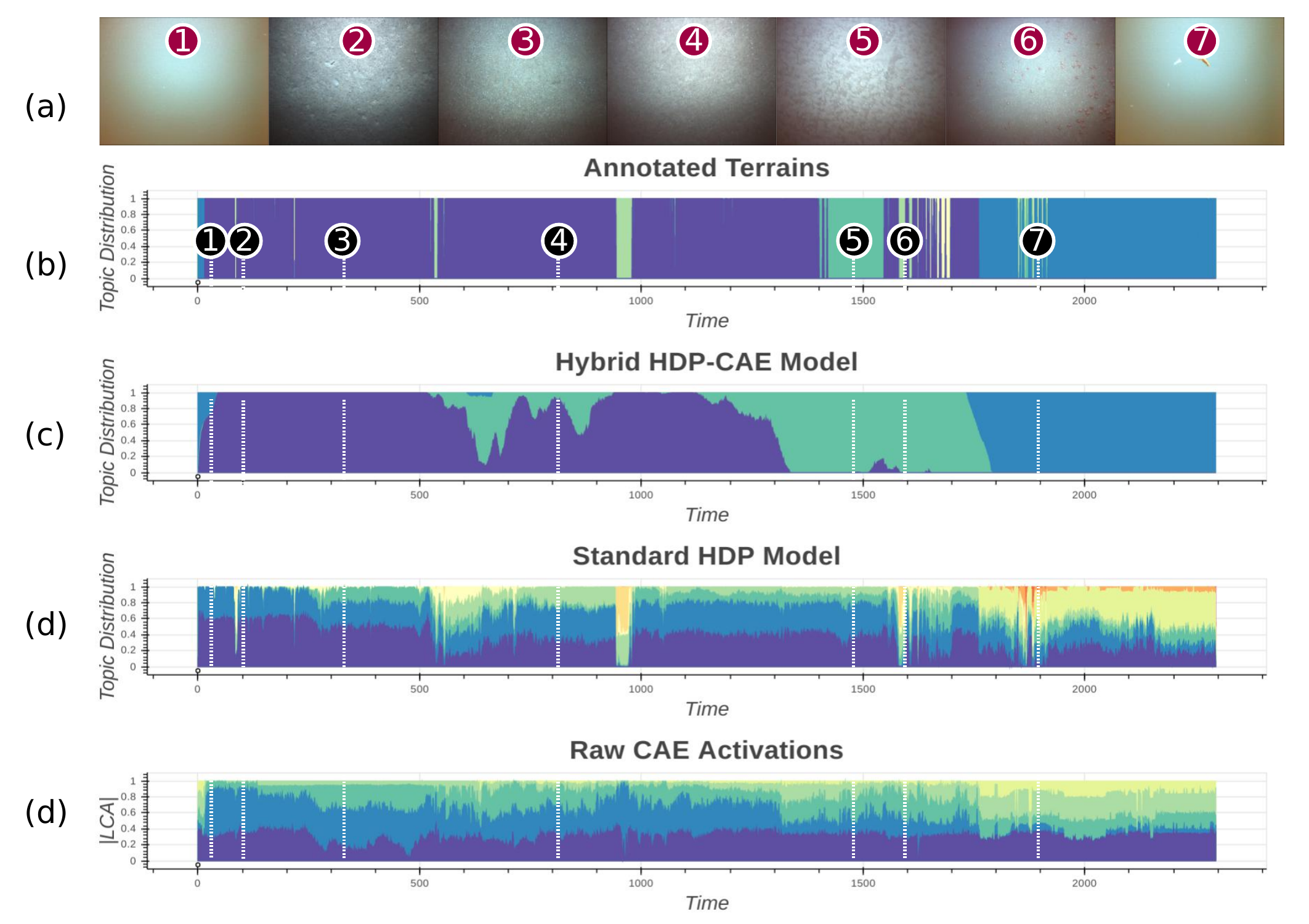}
  \caption{Results for two unsupervised topic models versus annotated labels in (b) for the Mission II dataset. Example images from the dataset are shown in (a). The hybrid HDP-CAE model (c) again outperforms the HDP model using standard image features (d). However, the hybrid model fails recognize some of the more transient topics, such as the crustacean swarm at (7). The learned feature representation is visualized as described in the text in (e).}    
\label{fig:missII}
\end{figure*}

\section{Experiments}
\label{sec:exp}

We evaluate our hybrid HDP-CAE model against a ROST HDP baseline using SURF \cite{Bay:ECCV:2006} and ORB \cite{RubleeE2011} features. The two models are visualized in Figure \ref{fig:overview}. We apply each model to image streams from two marine robot missions \cite{Pineda2016} and present experimental results.  

Mission I contains 1,117 images sampled every four seconds from a downwards facing camera mounted to the bottom of the robot. During Mission I, the robot passes over several seafloor terrains, including images of the water column, a rocky seafloor, and a porous sandy seafloor. This mission tests the model's ability to discover visually distinct terrain types. 

Mission II consists of 2,296 images sampled in a similar manner. Mission II contains mostly images of a sandy seafloor, interrupted several times by large, biologically interesting phenomena, such as crab congregations, seafloor carnage, and geothermal vents. In addition to terrain discovery, this mission also tests the model's ability to accurately identify anomalous images within a mission dataset.

To evaluate how well the image topic labels discovered by the unsupervised HDP-CAE model correspond to visually meaningful seafloor terrains, we hand-labeled each image in both missions with one of thirteen possible terrain labels, including: `water column', `sparse boulders', `smooth sand', `biological congregation', etc. These labels are used exclusively for model evaluation. 

We apply the HDP-ROST model described in Section \ref{sec:hdp} to the two mission datasets, using standard image features and CAE-derived LCA features for the standard HDP model and the hybrid HDP-CAE model respectively. New images are incorporated into the model in a streaming fashion; visual words are extracted from a new image and added to the model at regular intervals (200 ms). The ROST hyperparameters for Mission I and Mission II respectively were set to maximize mutual information between discovered topics and hand annotated labels: $\alpha = 0.1, 0.1; \beta = 25, 50; \gamma = 10^{-7}, 10^{-7}$. After Gibbs sampling, we have an approximation of the posterior over topics $z_i$ for an observed visual world $w_i$, $P(z_i | w_i = v)$.  The predicted topic label for each visual word is assigned as the \textit{maximum a posteriori} (MAP) topic label given by the posterior, and the predicted scene label $s_t$ for each image is calculated as the majority consensus of the visual words in the image.

Taking the MAP is a standard way of reducing a probabilistic distribution over a latent variable to a single point estimate. However, by using only the MAP topic label for each image, we are not allowing for important visual constructs to be represented by a mixture of topics. This approximation may be suitable for our experimental domain. The majority of images in our marine datasets consist of a homogeneous visual terrain, and an ideal topic model would discover topics rich enough to have nearly a one-to-one correspondence with semantically distinct visual constructs. Having to build a heuristic on top of a topic model to extract meaningful topics from mixtures of the discovered topics adds an unnecessary layer of complication to the model.

\section{Results}
To quantify the accuracy of the topic labels discovered by each of our models, we use normalized mutual information between the annotated topic distribution and the computed topic distribution. Mutual information captures the reduction in entropy of a random variable X after observing random variable Y (Eq. \ref{eq:mutualinf}). A normalized mutual information score of one indicates that X and Y are completely dependent (i.e. the discovered topics are completely correlated with the true labels), whereas a mutual information score of zero indicates independence (i.e. the discovered topics are unrelated to the true labels).
    
    \begin{equation} \label{eq:mutualinf}
    \begin{split}
        I(X, Y) & = H(X) - H(X|Y) \\
                & = \sum_{x,y} P(x,y) \log \frac{P(x,y)}{P(x)P(y)}
    \end{split}
    \end{equation}
Interactive visualizations of the experimental results presented here are available online\footnote{ http://warp.whoi.edu/iros2017/}.
    

\subsection{Mission I - Seafloor terrain discovery}
Mission I tests the model's ability to uncover topics corresponding to meaningful visual terrains. Table \ref{tab:mutualinf} shows that the topics discovered by the hybrid HDP-CAE model are highly predictive of the ground-truth seafloor terrains. The raw topic distribution (before MAP reduction) for the two models is plotted in Figure \ref{fig:missI} along with example images from the major terrain types. To generate the plots in Figure \ref{fig:missI}, visual words are extracted from an image at time $t$ and assigned a topic label $z_i$ as described in Section \ref{sec:approach}. The proportion of words in the image at time $t$ assigned to each topic label is shown on the y-axis, where different topics are represented by colors. Colors are unrelated across plots.

Although the hybrid HDP-CAE model differs from the human annotated terrains by, for example, not modeling the transient topic at (3), the major terrain transitions are captured faithfully. The rocky seafloor terrain that dominates at (4) and then partially appears again at (6) is assigned to the same topic. The moment that the robot first observes the seafloor through the water column in (3) is captured as a mixture of the `water column' topic (indigo) and the `rocky seafloor' topic (green).

\begin{table}[htb!]
\caption{Mutual information between \hspace{\textwidth} discovered topics
        and annotations}
\label{tab:mutualinf}
\centering
\begin{tabular}{ll|r}
     & Model & $I(X, Y)$ \\ \hline \\
    Mission I & Standard HDP &    0.185\\
     &   \textbf{Hybrid HDP-CAE} &   \textbf{0.535} \\
    Mission II &   Standard HDP &    0.123 \\
     &   \textbf{Hybrid HDP-CAE}  &   \textbf{0.441} \\ 
\end{tabular}
\end{table}
    
Despite its low mutual information scores, the standard HDP models does capture some of these transitions as changes in the topic distributions. However, it is not clear how to distill this information into meaningful topics. Because the hybrid HDP-CAE model uses much fewer, more abstract features, it is able to define topics that correspond more directly to useful visual phenomena. The discovered feature space is also visualized in Figure \ref{fig:missI}.

\begin{figure*}[thb!]
\centering
\includegraphics[scale=0.76]{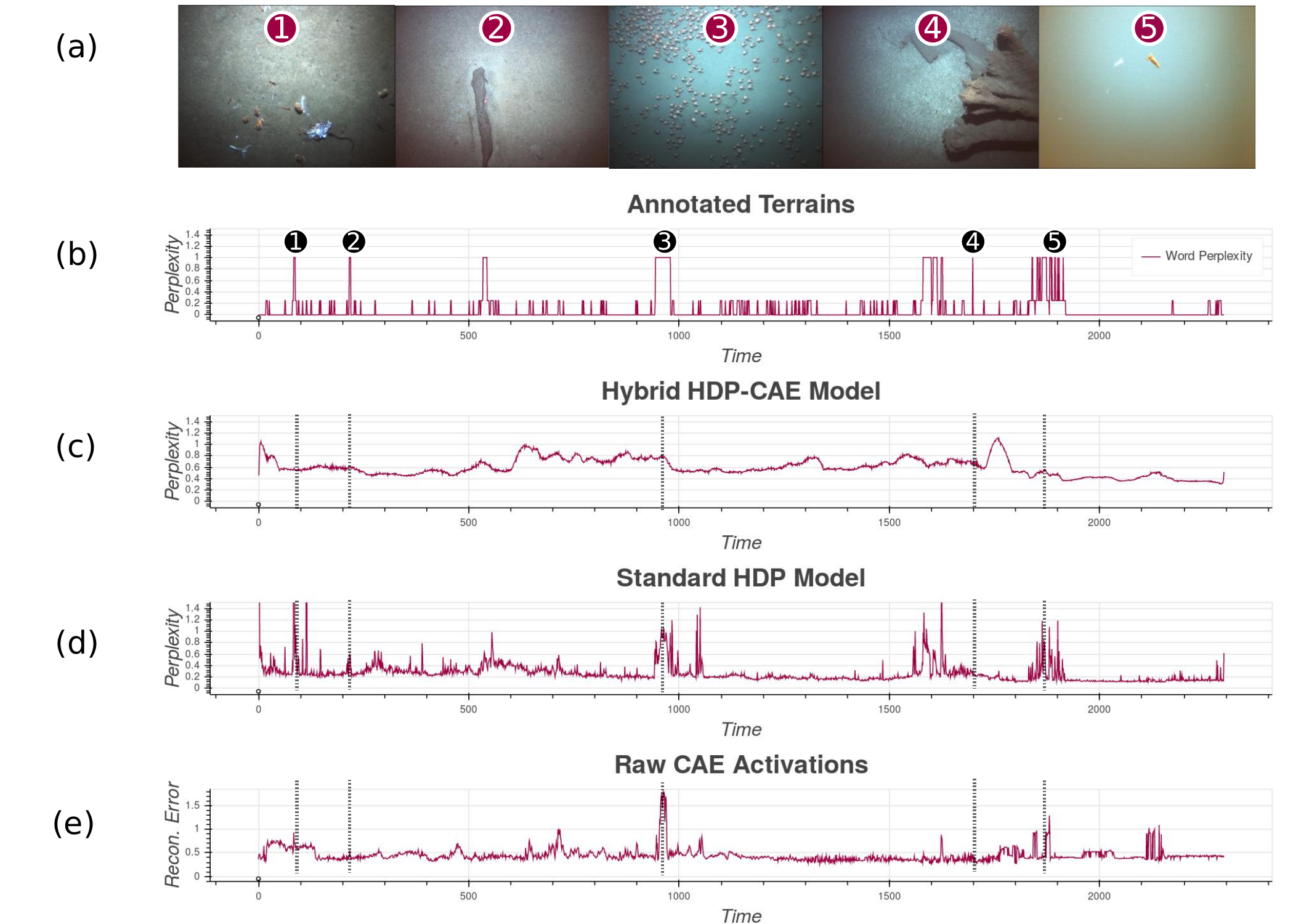}
    \caption{Correlation of perplexity score of the models (c,d,e) with annotated biological anomalies (b) for the Mission II dataset. Example images of biological anomalies are shown in (a). Each image in the dataset was labeled with high, medium, or low perplexity (b). Although all three models do have differential responses in areas of high perplexity, the HDP model using standard features (d) outperforms the hybrid HDP-CAE model (c) and raw reconstruction error from the CAE (e).}    
\label{fig:missII-perp}
\end{figure*}
    
\subsection{Mission II - Biological anomaly detection} 
\label{sec:missII}
Characterizing seafloor terrains is a vital task for an exploratory marine robot. Another complementary skill is the ability to identify images that are anomalous under the robot's current model of the world and flag these as interesting, one of the behaviors demonstrated by the standard HDP model presented in \cite{Girdhar2015}.  The Mission II dataset was designed to test both of these abilities.
    
By comparing mutual information between terrain labels, the hybrid HDP-CAE model again outperforms the baseline model. Figure \ref{fig:missII} shows the raw topic distribution for the two models, along with example images from the major terrain types. Mission II is a more visually homogeneous dataset, consisting almost entirely of sandy seafloor images. The hybrid HDP-CAE model discovers segmentations within the sandy seafloor topic that the human annotators do not, but otherwise captures major terrain transitions. However, the hybrid HDP-CAE model does not capture some of the more transient topics, such as the crustaceans at (7). 
    
To quantify this result further, we introduce the notion of perplexity. Because HDP is a probabilistic model, it is straightforward to quantify the average word perplexity (Eq. \ref{eq:perp}) of a new image $X_t$ under the model. \begin{equation}
\label{eq:perp}
Perplexity(X_t) = \exp\left( -\frac{\sum_{w \in W_t} \log p(w | X_t)}{|W_t|} \right)
\end{equation}
where the set $W_t$ consists of the visual words in $X_t$. High perplexity indicates that the image is not well modeled by the topic model, whereas low perplexity indicates an image that is well explained by the topic model. Figure \ref{fig:missII-perp} compares the perplexity response of each model when presented with biologically interesting images; the hybrid model does not have an obvious increase in perplexity when presented with the images of seafloor carnage (2), crab congregations (3), or submerged tree (4). 
    
To compute how well each model's perplexity score corresponds to some interesting visual phenomena, we annotated each image in the dataset as either high, medium, or low `interest' and computed the mutual information between the annotated and computed perplexity. Although perplexity scores do not necessarily correspond well with human intuition about semantic coherency \cite{Chang2009}, perplexity has been used successfully for anomaly detection in previous work \cite{Girdhar2015}. The CAE is not a probabilistic model, so there is no well-defined operation to compute perplexity. We instead use squared image reconstruction error as a proxy for perplexity.  The mean $\mu$ and standard deviation $\sigma^2$ of each model's perplexity distribution are used to bin the perplexity into low ($0 \leq x \leq \mu + \sigma^2$), medium($\mu + \sigma^2 \leq x \leq \mu + 2\sigma^2$), and high ($x > \mu + 2\sigma^2$) perplexity. The results of this analysis are shown in Table \ref{tab:perp}. Although all three models do have differential responses in areas of high annotated perplexity, the standard HDP model's perplexity has higher mutual information with annotated perplexity. We will address this discrepancy in the Discussion.
     
\begin{table}[htb!]
\caption{Mutual information between \hspace{\textwidth} perplexity and annotations}
\label{tab:perp}
\centering
\begin{tabular}{ll|r}
     & Model & $I(X, Y)$ \\ \hline \\
        Mission II &    \textbf{Standard HDP} &     \textbf{0.153}\\
     &   Hybrid HDP-CAE &  0.006 \\ 
     &   Raw CAE &     0.033\\ 
\end{tabular}
\end{table}
    
 \section{Discussion} \label{sec:disc}

The proposed hybrid HDP-CAE model significantly outperforms alternative models on the task of seafloor terrain discovery. The hybrid model, however, did not perform as well on the secondary task of anomaly detection, as quantified by image perplexity. Our hypothesis is that this limitation stems from the inherently imbalanced nature of anomalies within a dataset. In our anomaly detection experiments, the CAE was trained on over 2000 images of sandy seafloor and only 80 images of crab congregations. Many other anomalous events, such as the seafloor carnage in Figure \ref{fig:missII-perp}(a) appear for even shorter spans. Neural models have been shown to struggle when presented with imbalanced training data \cite{Chawla}. There may not be enough images of anomalous biological events for the CAE to learn a meaningful feature representation. Although the CAE image reconstruction error, plotted in Figure \ref{fig:missII-perp}(d), does capture the inability of the features to represent the anomalous images, our current hybrid HDP-CAE model does not incorporate this uncertainty within the topic modeling stage. An interesting extension of this work would be to use image reconstruction error directly as a metric of feature quality. Alternatively, there are methods within the machine learning community for dealing with imbalanced datasets that could improve CAE training. 

For the specific biological anomalies tested here, such as the crab congregations or seafloor carnage, it may be difficult to outperform standard image features, which are designed to detect areas of high image gradient. However, there are other reasons to prefer a CAE-based anomaly detector. Interesting anomalies may not always manifest themselves as complex visual structure; a smooth sandy seafloor is anomalous within a rocky mission. Standard image features may struggle to represent these visually simple anomalies. Additionally, CAE-based anomaly detectors can use reconstruction error to not only detect when an image is anomalous, but also which part of the image is particularly difficult to resolve. The ability to spatially localize anomalies could be very useful in robot scene understanding and real-time planning.

Another important extension to the hybrid HDP-CAE model presented in this work is the adaptation of convolutional feature discovery for realtime, streaming applications. Bayesian nonparametric models are well suited for the life-long learning required in streaming and robotics applications; this is one compelling reason to use them over purely neural models. However, for simplicity, the CAE-based feature discovery training in this work was done offline on complete datasets. Exploring methods for efficient, life-long training of convolutional models is an important area of future work for applying hybrid HDP-CAE models to realtime applications.
 
    
\section{Conclusions}
Bayesian topic models have achieved impressive performance by learning both model parameters and useful structure directly from data. However, these nonparametric models still fundamentally rely on predefined feature representations of data. We present a novel model that overcomes this limitation using convolutional autoencoders, allowing unsupervised discovery of \textit{both} a feature representation and thematic structure in image data. 

Our proposed hybrid model incorporates a convolutional autoencoder for data-driven feature discovery within a Bayesian topic modeling framework.  We apply this model to the problem of high-level scene understanding and mission visualization for exploratory marine robots. On complex mission datasets, the hybrid model discovers a rich latent visual structure that has over four times the mutual information with biologically meaningful seafloor terrains when compared to a Bayesian nonparametric topic model with standard, hand-designed features. This work defines a paradigm for including the ability of unsupervised neural models to discover useful, low-dimensional data representations within a Bayesian nonparametric topic modeling framework and demonstrates state of the art performance on a challenging problem from the marine robotics community.  Our future work will focus on adapting the convolutional model to run in real-time on a marine robot and improving the model's ability to detect visual anomalies in an image dataset.

\section*{Acknowledgements}
This work was supported in part by an NSF Graduate Research Fellowship Program award and The John P. Chase Memorial Endowed Fund.

\bibliographystyle{IEEEtran}
\bibliography{IEEEabrv,girdhar,iros_genevieve}
\end{document}